\documentclass[10pt,twocolumn,letterpaper]{article}

\usepackage{iccv}

\usepackage{float}
\usepackage{times}
\usepackage[dvipdfmx]{graphicx}
\usepackage{amsmath}
\usepackage{amssymb}
\usepackage{multirow}
\usepackage{caption}

\DeclareMathOperator*{\argmin}{arg\,min}
\usepackage{subfiles}
\usepackage[ruled,vlined]{algorithm2e}
\usepackage{subcaption}
\usepackage{tablefootnote}

\usepackage[pagebackref=true,colorlinks]{hyperref}
\usepackage{cleveref}

\iccvfinalcopy 

\def\iccvPaperID{3480} 

\ificcvfinal\fi

\begin{document}

\title{RePOSE: Fast 6D Object Pose Refinement via Deep Texture Rendering}

\author{
Shun Iwase$^1$ \qquad 
Xingyu Liu$^1$ \qquad 
Rawal Khirodkar$^1$ \qquad
Rio Yokota$^2$ \qquad
Kris M. Kitani$^1$
\\$^1$Carnegie Mellon University \qquad $^2$Tokyo Institute of Technology \\
}

%
%
%
%

\maketitle
\ificcvfinal\thispagestyle{empty}\fi

\begin{abstract}
  We present RePOSE, a fast iterative refinement method for 6D object pose estimation. Prior methods perform refinement by feeding zoomed-in input and rendered RGB images into a CNN and directly regressing an update of a refined pose. Their runtime is slow due to the computational cost of CNN, which is especially prominent in multiple-object pose refinement. To overcome this problem, RePOSE leverages image rendering for fast feature extraction using a 3D model with a learnable texture. We call this deep texture rendering, which uses a shallow multi-layer perceptron to directly regress a view-invariant image representation of an object. Furthermore, we utilize differentiable Levenberg-Marquardt (LM) optimization to refine a pose fast and accurately by minimizing the distance between the input and rendered image representations without the need of zooming in. These image representations are trained such that differentiable LM optimization converges within few iterations. 
Consequently, RePOSE runs at $92$ FPS and achieves state-of-the-art accuracy of $51.6$\% on the Occlusion LineMOD dataset - a $4.1$\% absolute improvement over the prior art, and comparable result on the YCB-Video dataset with a much faster runtime. The code is available at \href{https://github.com/sh8/repose}{https://github.com/sh8/repose}.
\end{abstract}


\begin{figure}[t]
\hspace*{-0.2cm}
\centering
\includegraphics[scale=0.33]{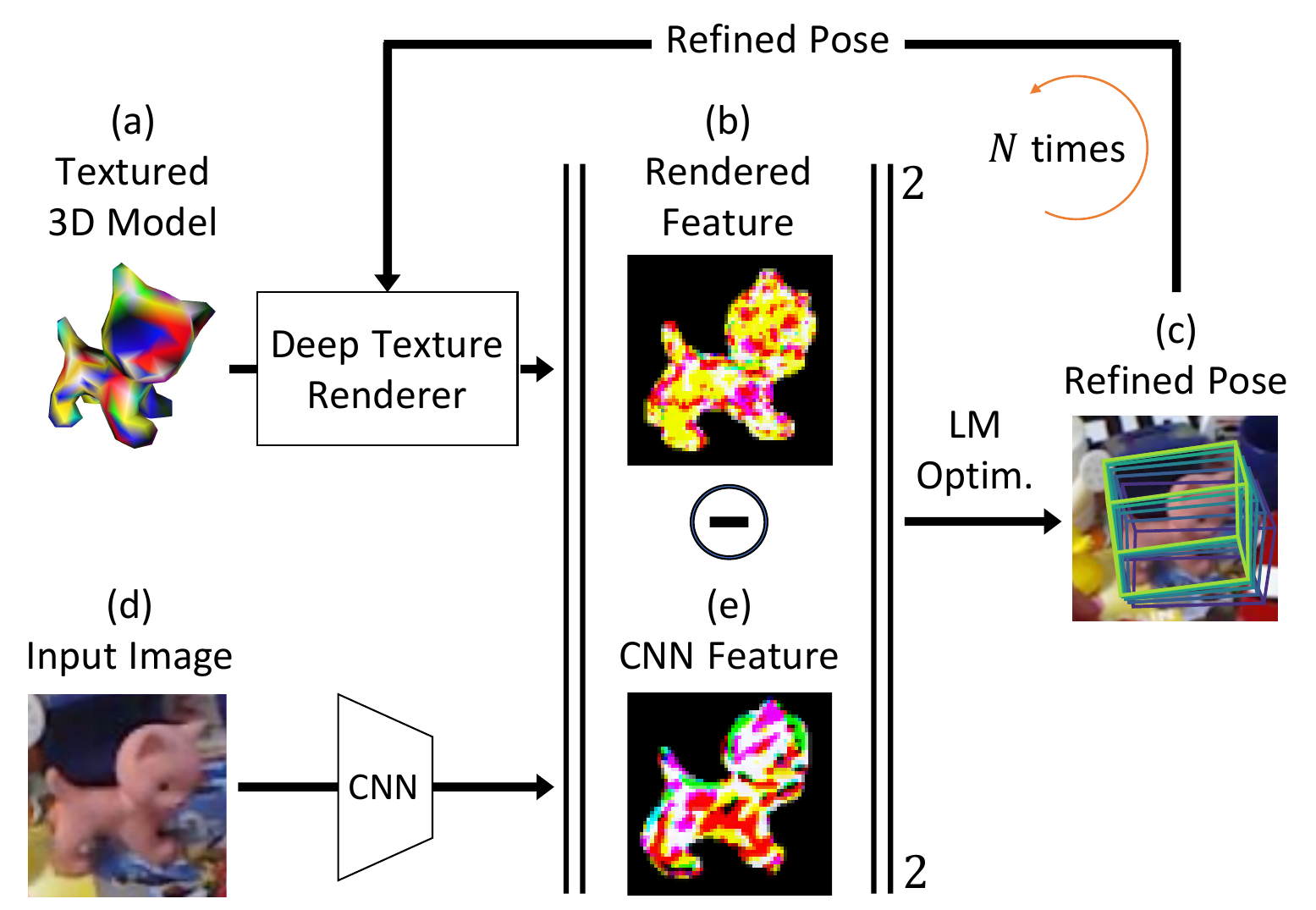}
\caption{\textbf{RePOSE Framework:} (a) 3D model with deep texture is projected to obtain (b) the rendered image representation with the deep texture renderer. (c) The pose is refined iteratively by minimizing the projection error of the rendered image representation and (e) the CNN feature extracted from (d) the input image via Levenberg-Marquardt (LM) optimization.
}
\label{fig:pipeline}
\end{figure}

\section{Introduction}
\label{sec:introduction}

In many applications of 6D object pose estimation like robotic grasping and augmented reality (AR), fast runtime is critical.
State-of-the-art 6D object pose estimation methods~\cite{labbe2020,Zakharov_2019_ICCV,DBLP:journals/corr/RadL17} demonstrate that iterative 6D object pose refinement improves the accuracy largely. Nevertheless, since recent 6D object pose refinement methods~\cite{li2018deepim,labbe2020} directly regress an update of a pose to align a zoomed-in input image of an object against a template image (\eg, 3D rendering of that object) using a Convolutional Neural Network (CNN), we presume that the CNN's computational cost of zoomed-in inputs can be a bottleneck toward the real-time 6D object pose estimation. 

We have mainly two choices of refinement strategies. As described, the former one is CNN-based direct regression, which generally requires large computational cost. The latter one is a classical non-linear optimization~\cite{10.1007/BFb0067700} which iteratively updates a pose by minimizing the photometric error between input and template images. Their runtime per iteration is quite fast. Since the photometric error explicitly considers each pixel, they can obtain enough details for accurate optimization without the need of zooming in. However, they can fail under diverse illumination or gross pose differences. Although non-linear least squares methods such as inverse compositional image alignment~\cite{10.1023/B:VISI.0000011205.11775.fd,10.5555/1623264.1623280} or active appearance models~\cite{10.1007/BFb0054760,Matthews-2003-8630} are extremely efficient, straightforward implementations of such methods can be unstable under significant illumination or pose changes. In addition, their runtime can be slower if many iterations are performed until convergence.

We leverage and improve the latter method to realize both quick and accurate refinement. In this paper, we propose RePOSE, a new feature-based non-linear optimization framework for 6D object pose refinement. The main technical insight presented in this work is that one can learn an image feature representation which is both robust for alignment and fast to compute. As stated earlier, the main impediment of CNN-based refinement methods is that the deep feature must be extracted during the refinement process iteratively. To remove this, we show that it is possible to directly render deep features using a simple graphics render. The rendering process decouples the shape of the object from the texture. At the time of rendering, texture is mapped to the 3D shape and then projected as a 2D image. Instead of mapping an RGB valued texture to the object, we can alternatively render a deep feature texture. Then, the rendered object can be directly aligned to the deep features of the input image. By retaining the deep feature representation during rendering, the pose alignment is robust and the refinement process becomes very efficient.

RePOSE refines an object pose by minimizing the distance between the deep features of the input and rendered images. Since the input image is fixed during iterative refinement, its feature is only computed once using a CNN. In contrast, the deep feature of the template image are directly generated using a simple computer graphics renderer. The rendering process takes less than a millisecond which greatly increases the speed of the iterative refinement process. The deep feature representation is learned such that nonlinear optimization can be easily performed through a differentiable LM optimization network ~\cite{10.1007/BFb0067700}. We experimentally found 5 iterations are enough to converge, which contributes to fast 6D object pose refinement.

RePOSE has several practical advantages over recent CNN-based regression methods: 1) RePOSE can be exceptionally fast. --- In the case of 1 iteration, RePOSE runs at 181 FPS for 5 objects and 244 FPS for 1 object, 2) RePOSE is data efficient. --- Since RePOSE considers projective geometry explicitly, there is no need to learn the mapping of the deep feature into an object’s pose from training data. In our experiments, we show that RePOSE achieves better or comparable performance with much fewer number of training images than prior methods, and 3) RePOSE does not request RGB textures of a 3D model. --- It has been known that RGB texture scanning has troubles with metalic, dark-colored, or transparent objects even with the latest 3D scanner~\cite{EinScanPro}. We believe that the requirement of RGB textures of recent CNN-based regression methods~\cite{li2018deepim,labbe2020} makes the implementation in the real world more challenging.

We evaluate RePOSE on three popular 6D object estimation datasets - LineMOD \cite{linemod}, the challenging Occlusion LineMOD \cite{10.1007/978-3-319-10605-2_35}, and YCB-Video \cite{xiang2018posecnn}. RePOSE sets a new state of the art on the Occlusion LineMOD (51.6\%) \cite{10.1007/978-3-319-10605-2_35} dataset and achieves comparable performance on the other datasets with much faster speed ($80$ to $92$ FPS with 5 iterations). Additionally, we perform ablations to validate the effectiveness of our proposed methods.

\vspace{0.5cm}
\begin{figure*}
\begin{center}
\includegraphics[width=0.95\linewidth]{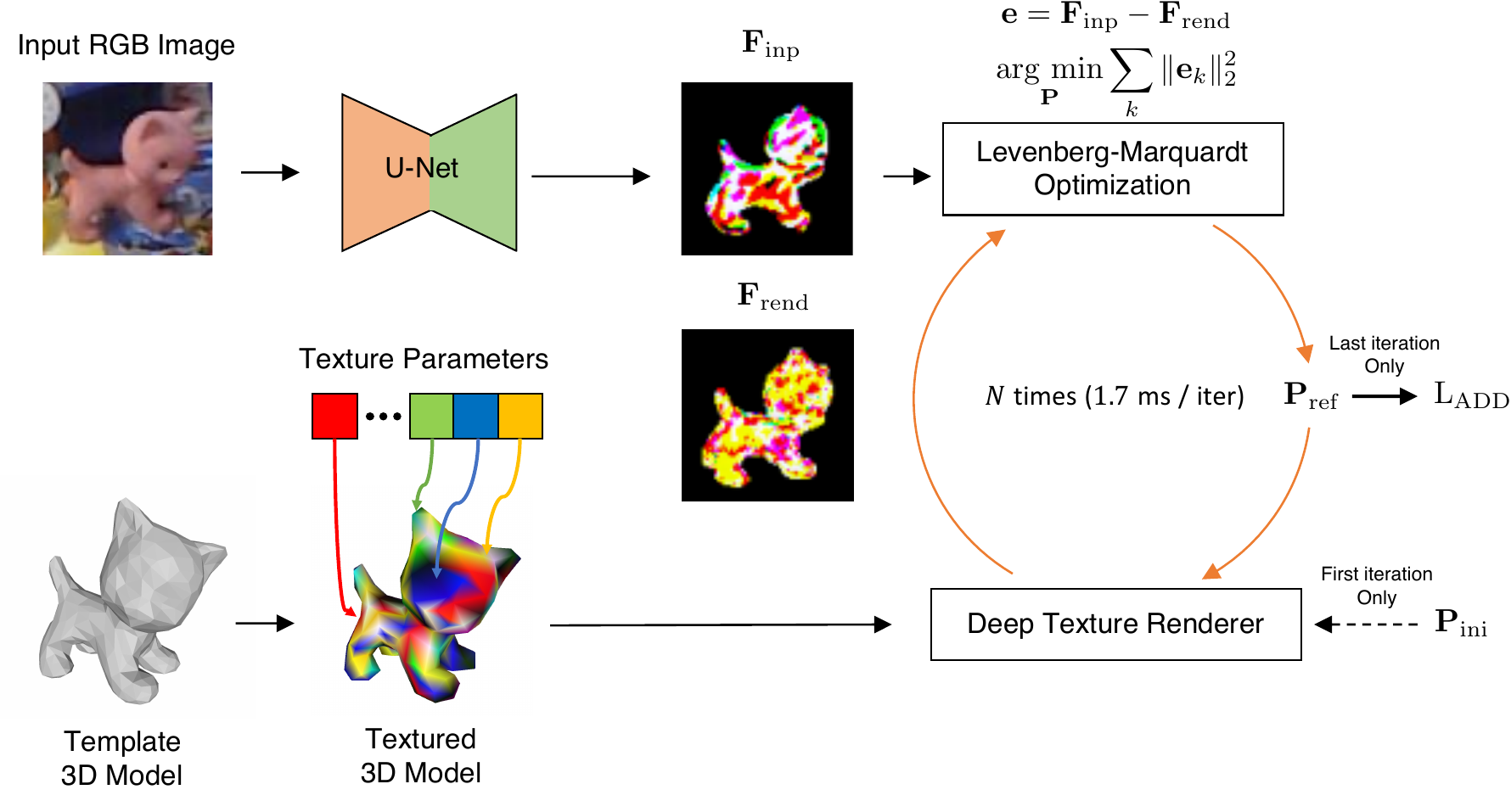}
\end{center}
\caption{Overview of the RePOSE refinement network. Given an input image $\mathbf{I}$ and the template 3D model $M$ with deep textures, U-Net and deep texture renderer output features $\mathbf{F}_\text{inp}$ and $\mathbf{F}_\text{rend}$ respectively. We use Levenberg-Marquardt optimization \cite{10.1007/BFb0067700} to obtain the refined pose $\mathbf{P}_{ref}$. The refined pose $\mathbf{P}_{ref}$ after $N$ iterations is used to compute the loss $L_{ADD(-S)}$. The pre-trained encoder of the initial pose estimator is used. The decoder of U-Net and deep textures (seed parameters, and fc layers) are trained to minimize $L_\text{ADD(-S)}$ and $L_\text{diff}$.}
\label{fig:overview}
\end{figure*}

\section{Related Work}
\label{sec:related_work}
\paragraph{Two-stage pose estimation methods}

Recently, Oberweger~\cite{Oberweger_2018_ECCV}, PVNet \cite{peng2019pvnet}, DPOD \cite{Zakharov_2019_ICCV}, and HybridPose \cite{Song_2020_CVPR} have shown excellent performance on 6D object pose estimation using a two-stage pipeline to estimate a pose: (i) estimating a 2D representation (\eg keypoints, dense correspondences, edge vectors, symmetry correspondences), (ii) PnP algorithm~\cite{10.1007/s11263-008-0152-6,BPnP2020} for pose estimation. DOPE~\cite{tremblay2018corl:dope} and BB8~\cite{DBLP:journals/corr/RadL17} estimate the corners of the 3D bounding box and run a PnP algorithm. Instead of regarding the corners as keypoints, PVNet~\cite{peng2019pvnet} places the keypoints on the object surface via the farthest point sampling algorithm. PVNet also shows that their proposed voting-based keypoint detection algorithm is effective especially for occluded objects. HybridPose~\cite{Song_2020_CVPR} uses multiple 2D representations including keypoints, edge vectors, and symmetry correspondences and demonstrates superior performance through constraint optimization.
DPOD~\cite{Zakharov_2019_ICCV} takes advantage of the dense correspondences using a UV map as a 2D representation. However, since the PnP algorithm is sensitive to small errors in the 2D representation, it is still challenging to estimate the object pose especially under occlusion. RePOSE adopts PVNet~\cite{peng2019pvnet} as the initial pose estimator using the official implementation.

\paragraph{Pose refinement networks}
Recent works ~\cite{xiang2018posecnn,Sundermeyer_2018_ECCV,Zakharov_2019_ICCV,Song_2020_CVPR,li2018deepim} have demonstrated that using a pose refinement network after the initial pose estimator is effective for 6D object pose estimation. For practical applications, the runtime of the pose refinement network is crucial. PoseCNN~\cite{xiang2018posecnn} and AAE~\cite{Sundermeyer_2018_ECCV} incorporates an ICP algorithm~\cite{Zhang2014} using depth information to refine the pose with a runtime of around $200$ ms. SSD6D~\cite{Kehl_2017_ICCV} and HybridPose~\cite{Song_2020_CVPR} proposed to refine the pose by optimizing a modification of reprojection error. DeepIM~\cite{li2018deepim}, DPOD~\cite{Zakharov_2019_ICCV}, and CosyPose~\cite{labbe2020} introduce a CNN-based refinement regression network using the zoomed-in input image and a rendered object image. Their methods require a high-quality texture map of a 3D model to compare the images. However, it is still challenging to obtain accurate texture scans of metalic, dark-colored, or transparent objects. NeMO~\cite{wang2021nemo} proposes a pose refinement method using the standard differentiable rendering and learning the texture of a 3D model via contrastive loss. However, gradient descent is used for optimization, hence, it takes more than $8$s for inference and is not fast enough for real-time applications. 

\paragraph{Non-linear least squares optimization}
Non-linear least squares optimization is widely used in machine learning. In computer vision, it is often utilized to find an optimal pose which minimizes the reprojection error or photometric error \cite{Alismail-2016-5532,7219438,7780814}. Recently some works~\cite{Tang2019,gnnet,Clark_2018_ECCV} incorporate non-linear least squares algorithms like Gauss-Newton and Levenberg-Marquardt~\cite{10.1007/BFb0067700} into a deep learning network for efficient feature optimization in VisualSLAM\@. RePOSE is inspired by similar formulation as in ~\cite{Tang2019}.

\section{RePOSE: Fast 6D Object Pose Refinement}
\label{sec:method}
Given an input image $\mathbf{I}$ with a ground-truth object pose $\mathbf{P}_\text{gt}$ and the template 3D model $\mathcal{M}$, RePOSE predicts pose $\mathbf{\hat{P}}$ of model $M$ which matches $\mathbf{P}_\text{gt}$ in $\mathbf{I}$. We extract a feature $\mathbf{F}_\text{inp}$ from image $\mathbf{I}$ using a CNN $\Phi$ \ie $\mathbf{F}_\text{inp} = \Phi(\mathbf{I})$. RePOSE then refines the initial pose estimate $\mathbf{P}_\text{ini} = \Omega(\mathbf{I})$ where $\Omega$ is any pose estimation method like PVNet~\cite{peng2019pvnet} and PoseCNN~\cite{xiang2018posecnn} in real time using differentiable Levenberg–Marquardt (LM) optimization \cite{10.1007/BFb0067700}. RePOSE renders the template 3D model with learnable deep textures in pose $\mathbf{P}$ to extract feature $\mathbf{F}_\text{rend}$. The pose refinement is performed by minimizing the distance between $\mathbf{F}_\text{inp}$ and $\mathbf{F}_\text{rend}$. We now describe in detail (1) $\mathbf{F}_\text{inp}$ extraction, (2) $\mathbf{F}_\text{rend}$ extraction and finally (3) the pose refinement using LM optimization.

\subsection{Feature Extraction of an Input Image $\mathbf{F}_\text{inp}$} 
\label{sec:deep_feet_ext}

We adopt a U-Net \cite{RFB15a} architecture for the CNN $\Phi$. The decoder outputs a deep feature map for every pixel in $\mathbf{I}$. The per-pixel feature $\mathbf{F}_\text{inp} \in \mathbb{R}^{w \times h \times d}$ is extracted by the decoder. Figure \ref{fig:pipeline} (b) provides a visual illustration of $\mathbf{F}_\text{inp}$ extracted from the input image $\mathbf{I}$. Note that the channel depth $d$ is a flexible parameter but we found $d=3$ to be optimal. The pre-trained weights of PVNet~\cite{peng2019pvnet} or PoseCNN~\cite{xiang2018posecnn} are used for the encoder and only the decoder is trained while training RePOSE. 

\subsection{Template 3D Model Rendering $\mathbf{F}_\text{rend}$}
\label{sec:deep_feet_rend}

The template 3D model $\mathcal{M}$ with pose $\mathbf{P} = \{ \mathbf{R}$,  $\mathbf{t} \}$ where $\mathbf{R}$ is 3D rotation and $\mathbf{t}$ is 3D translation, is projected to 2D to render the feature $\mathbf{F}_\text{rend}$. Let the template 3D model $\mathcal{M}=\{ \mathcal{V},\mathcal{C}, \mathcal{F} \}$ be represented by a triangular watertight mesh consisting of $N$ vertices $\mathcal{V} = \{ V_n \}^N_{n=1}$ where $V_n \in \mathbb{R}^{3}$, faces $\mathcal{F}$ and deep textures $\mathcal{C}$. $V_n$ is the 3D coordinate of the vertex in the coordinate system centered on the object. Each vertex $V_n$ has a corresponding vertex learnable texture $\mathbf{C}_n \in \mathbb{R}^{d}$, $\mathcal{C} = \{ \mathbf{C}_n \}^N_{n=1}$, which is learned. Note that the dimensions of the vertex learnable texture $d$ must match depth dimension of input image feature $\mathbf{F}_\text{inp}$ so that they can be compared during alignment.

\begin{figure}[t]
  \centering
  \includegraphics[scale=0.3]{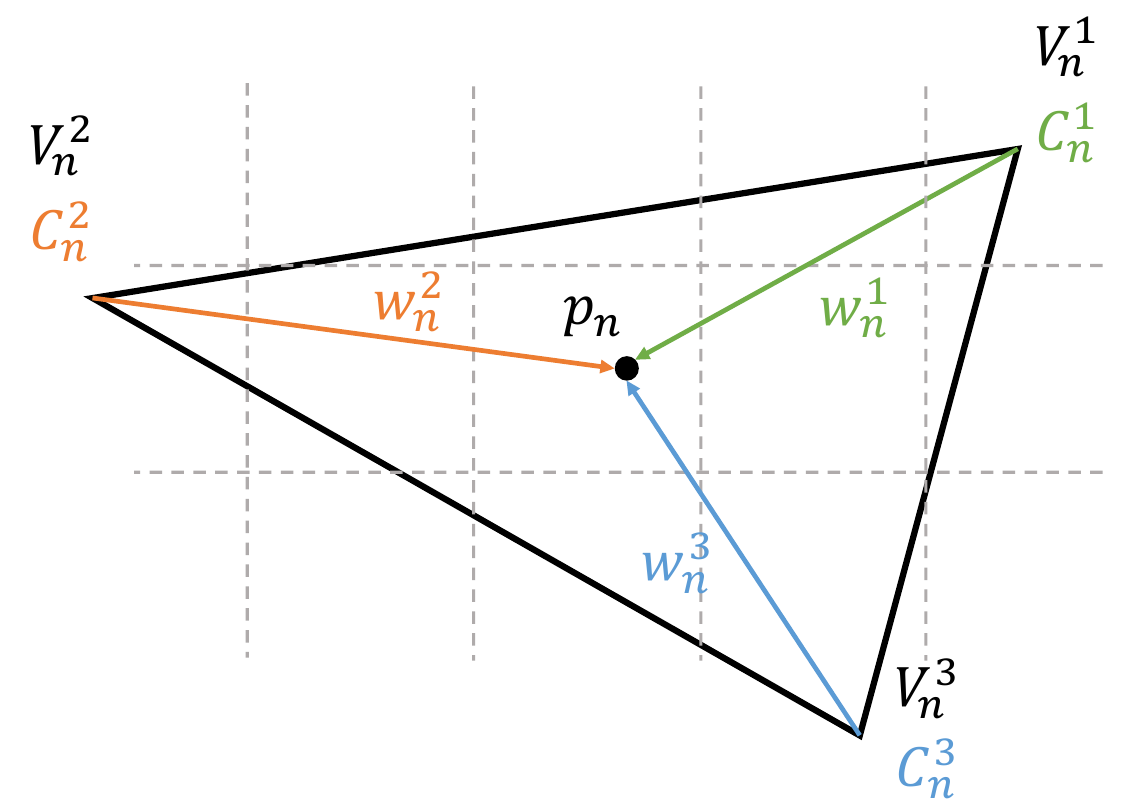}
  \caption{Rasterization of deep textures into a pixel $p_n$ as the weighted sum of $\mathbf{C}^i_n$ using $w^i_n$ as weights in the barycentric coordinate system, $\sum_i^3 w^i_n = 1$.}
  \label{fig:rasterize}
\end{figure}

\begin{figure*}
\begin{center}
\hspace*{-0.5cm}
\begin{tabular}{cccccccc}
  \begin{minipage}{0.10\linewidth}
    \includegraphics[scale=0.67]{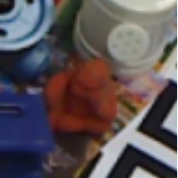}
  \end{minipage} &
  \begin{minipage}{0.10\linewidth}
    \includegraphics[scale=0.67]{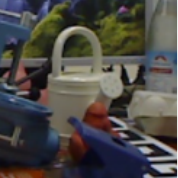}
  \end{minipage} &
  \begin{minipage}{0.10\linewidth}
    \includegraphics[scale=0.67]{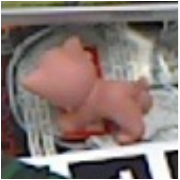}
  \end{minipage} &
  \begin{minipage}{0.10\linewidth}
    \includegraphics[scale=0.67]{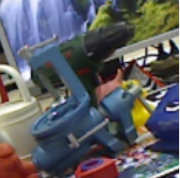}
  \end{minipage} &
  \begin{minipage}{0.10\linewidth}
    \includegraphics[scale=0.67]{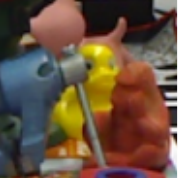}
  \end{minipage} &
  \begin{minipage}{0.10\linewidth}
    \includegraphics[scale=0.67]{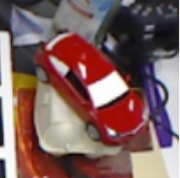}
  \end{minipage} & 
  \begin{minipage}{0.10\linewidth}
    \includegraphics[scale=0.67]{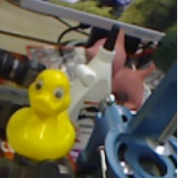}
  \end{minipage} &
  \begin{minipage}{0.10\linewidth}
    \includegraphics[scale=0.67]{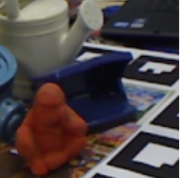}
  \end{minipage} \\ \\
    \begin{minipage}{0.10\linewidth}
    \includegraphics[scale=0.67]{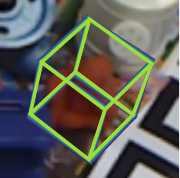}
  \end{minipage} &
  \begin{minipage}{0.10\linewidth}
    \includegraphics[scale=0.67]{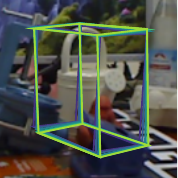}
  \end{minipage} &
  \begin{minipage}{0.10\linewidth}
    \includegraphics[scale=0.67]{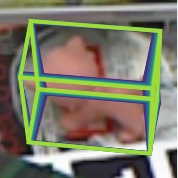}
  \end{minipage} &
  \begin{minipage}{0.10\linewidth}
    \includegraphics[scale=0.67]{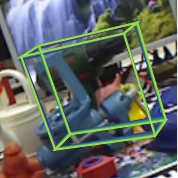}
  \end{minipage} &
  \begin{minipage}{0.10\linewidth}
    \includegraphics[scale=0.67]{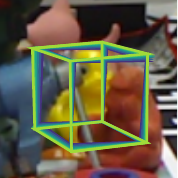}
  \end{minipage} &
  \begin{minipage}{0.10\linewidth}
    \includegraphics[scale=0.67]{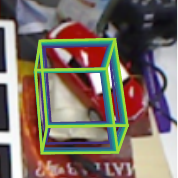}
  \end{minipage} & 
  \begin{minipage}{0.10\linewidth}
    \includegraphics[scale=0.67]{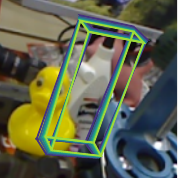}
  \end{minipage} &
  \begin{minipage}{0.10\linewidth}
    \includegraphics[scale=0.67]{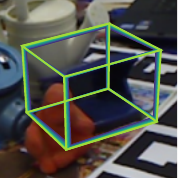}
  \end{minipage} \\ \\
   \begin{minipage}{0.10\linewidth}
    \includegraphics[scale=0.67]{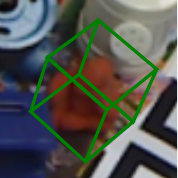}
    \caption*{{(a) ape}}
  \end{minipage} &
  \begin{minipage}{0.10\linewidth}
    \includegraphics[scale=0.67]{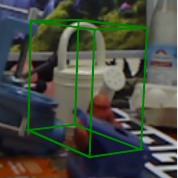}
    \caption*{{(b) can}}   
  \end{minipage} &
  \begin{minipage}{0.10\linewidth}
    \includegraphics[scale=0.67]{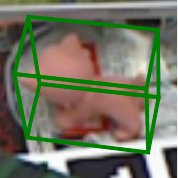}
    \caption*{{(c) cat}}   
  \end{minipage} &
  \begin{minipage}{0.10\linewidth}
    \includegraphics[scale=0.67]{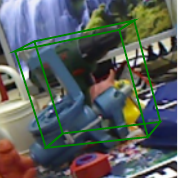}
    \caption*{{(d) driller}}   
  \end{minipage} &
  \begin{minipage}{0.10\linewidth}
    \includegraphics[scale=0.67]{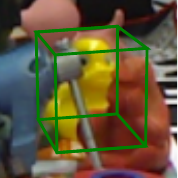}
    \caption*{{(e) duck}}   
  \end{minipage} &
  \begin{minipage}{0.10\linewidth}
    \includegraphics[scale=0.67]{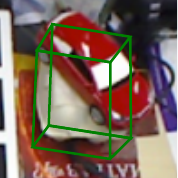}
    \caption*{{(f) eggbox}}   
  \end{minipage} & 
  \begin{minipage}{0.10\linewidth}
    \includegraphics[scale=0.67]{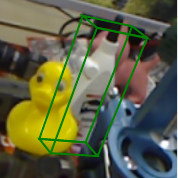}
    \caption*{{(g) glue}}      
  \end{minipage} &
  \begin{minipage}{0.10\linewidth}
    \includegraphics[scale=0.67]{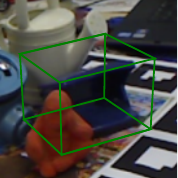}
    \caption*{{(h) hole.}}         
  \end{minipage}  
\end{tabular}
\end{center}
\caption{Example results on the Occlusion LineMOD dataset \cite{10.1007/978-3-319-10605-2_35}. We show an input RGB image, refined poses, and ground-truth pose from the top to bottom. The color of 3D bounding boxes are changed from purple to lightgreen as optimization progresses.}
\label{fig:short}
\end{figure*}

RePOSE projects the 3D mesh onto to the image plane using a pinhole camera projection function $\pi$ (homogeneous to inhomogeneous coordinate conversion). Specifically, we map the vertex $\mathbf{V}_n$ to $\mathbf{v} \in \mathbb{R}^{2}$ using eq \ref{eq:projection}. 
\begin{equation}
   \mathbf{v}_n = \pi \left( \mathbf{V}_n \mathbf{R}^{\top} + \mathbf{t}^{\top} \right) ~~ \forall ~ n
\label{eq:projection}
\end{equation}

The vertex deep textures $\mathbf{C}_n \in \mathbb{R}^{3}$ are learnable and computed using a 2-layer fully-connected network. The deep texture at each pixel is calculated by rasterization using the deep textures $\mathbf{C}_n$ in barycentric coordinates $w$ as shown in \Cref{fig:rasterize}. This operation can be parallelized using a GPU. Our custom implementation of the \cite{kato2018renderer}'s renderer takes less than $1$ ms to render $\mathbf{F}_\text{rend}$. $\mathbf{F}_{\text{rend}}(x,y)$ at a pixel location $(x,y)$ is computed as follows:
\begin{equation}
  \mathbf{F}_{\text{rend}}(x,y) = \sum_{i=1}^{3} w_n^{i} C_n^{i}
  \label{eq:rasterizing}
\end{equation}
where the triangular face index $n$ corresponding to the pixel $p_n$ at $(x, y)$ is found by ray tracing and $w_i$ is the normalized barycentric weight corresponding to the coordinates $(x, y)$ inside the triangle (\Cref{fig:rasterize}). Simply put, the rendered deep feature $\mathbf{F}_{\text{rend}}(x,y)$ is a linear combination of deep textures of the three projected vertices.

$\mathbf{F}_{\text{rend}}$ is end-to-end learnable by backpropagation. The gradient of $\mathbf{F}_{\text{rend}}$ with respect to the three deep textures of the triangle $\{C_n^i\}_{i=1}^3$ is as follows:
\begin{equation}
  \frac{\partial{ \mathbf{F}_\text{rend}(x,y) }}{ \partial{C_n^i}} = w_n^i.
  \label{eq:rasterizing_deriv}
\end{equation}

Note that $\mathbf{F}_\text{rend}$ is the output of a non-linear function $\Psi$ of the template 3D model $\mathcal{M}$ and its pose $\mathbf{P}$, \ie, $\mathbf{F}_\text{rend} = \Psi (\mathbf{P}, \mathcal{M})$ where $\Psi$ is the deep texture renderer (\Cref{fig:overview}).

\subsection{Levenberg-Marquardt (LM) Optimization}
After computing $\mathbf{F}_\text{inp}$ (\Cref{sec:deep_feet_ext}) and $\mathbf{F}_\text{rend}$ (\Cref{sec:deep_feet_rend}), the optimal pose $\mathbf{\hat{P}}$ is calculated by minimizing the following objective function:
\begin{align}
    \mathbf{e} &= \text{vec}(\mathbf{F}_\text{inp}) - \text{vec}(\mathbf{F}_\text{rend}), \\
    \mathbf{\hat{P}} &=\argmin_{\mathbf{P}} \sum_{k} ||e_k||^2_2,
  \label{eq:error_lm}
\end{align}
where $e_k$ denotes the $k^{\text{th}}$ element of the error $\mathbf{e} \in \mathbb{R}^{whd}$ and is the element-wise difference between the flattened values of $\mathbf{F}_\text{inp}$ and $\mathbf{F}_\text{rend}$. To perform optimization efficiently, we only use the error $\mathbf{e}$ in the pixel where the mask of $\mathbf{F}_\text{rend}$ exists.

We solve this non-linear least squares problem using the iterative Levenberg-Marquardt (LM) algorithm. The update rule for the pose $\mathbf{P}$ is as follows:
\begin{align}
  \Delta \mathbf{P} 
  &= (\mathbf{J}^T \left(\mathbf{e}\right) \mathbf{J} + \lambda \mathbf{I})^{-1} \mathbf{J}^T \left(\mathbf{e}\right) \mathbf{e},
  \label{eq:lavenberg}
\\
\mathbf{P}_{i+1} &= \mathbf{P}_i + \Delta \mathbf{P},
\label{eq:lavenberg_update}
\end{align}
where $\mathbf{J}$ is the Jacobian of the objective with respect to the pose $\mathbf{P}$, and $\lambda$ is a learnable step size.

The Jacobian $\mathbf{J}$ can be decomposed as:
\begin{equation}
  \mathbf{J} =
  \cfrac{\partial \mathbf{F}_\text{rend}}{\partial \mathbf{P}}
  =
  \cfrac{\partial \mathbf{F}_\text{rend}}{\partial \mathbf{x}}
  \cfrac{\partial \mathbf{x}}{\partial \mathbf{P}}
  \label{eq:jacobian}
\end{equation}
where $\mathbf{x}$ is a vector of all 2D image coordinate. We compute $\frac{\partial \mathbf{F}_\text{rend}}{\partial \mathbf{x}}$ using a finite difference approximation and $\frac{\partial \mathbf{x}}{\partial \mathbf{P}}$ is computed analytically. Please refer supplemental for the details. 

We minimize a loss function $\mathcal{L}_\text{ADD(-S)}$ based on the ADD(-S) score:
\begin{equation}
  \mathcal{L}_\text{ADD(-S)} = S_\text{ADD(-S)} (\mathbf{P},\mathbf{P}_\text{gt})
\end{equation}
where $S$ is the function used to calculate the distance used in the ADD(-S) score.
Additionally, we also minimize a loss function $\mathcal{L}_\text{diff}$ which ensures the value of the objective function is minimized when the pose $\mathbf{P}$ is equal to $\mathbf{P}_\text{gt}$:
\begin{align}
    \mathbf{d} 
    &= \text{vec}(\mathbf{F}_\text{inp}) - \text{vec}(\Psi(\mathbf{P}_\text{gt}, \mathcal{M})), \\
    \mathcal{L}_\text{diff} 
    &=  \sum_k ||d_k||^2_2.
\end{align}

The minimization of these two loss functions through LM optimization allows our refinement network to learn representations of the input image as well as the rendered object image, which helps in predicting the optimal pose.
\begin{equation}
\mathcal{L} = \mathcal{L}_\text{ADD(-S)} + \alpha \mathcal{L}_\text{diff}
\end{equation}
where $\alpha$ is a hyperparameter.

We show the RePOSE framework in \Cref{alg:repose}.
Note, all the operations inside the LM optimization (\Cref{eq:lavenberg,eq:lavenberg_update}) are differentiable allowing us to learn deep textures $\mathcal{C}$ and $\Phi$ using backpropagation. 

\begin{algorithm}[t]
\SetAlCapNameFnt{\small}
\SetAlCapFnt{\small}
\small{
\caption{RePOSE Training}\label{alg:repose}
$\mathcal{V} = \textsc{VerticesOf3DModel()}$; \\
$\mathcal{F} = \textsc{FacesOf3DModel()}$; \\
$\mathcal{C} = \textsc{InitializeTextureParameters()}$; \\
\# Iterate over Training Data \\
\For{$\mathbf{P}_\text{ini}, \mathbf{P}_\text{gt}, \mathbf{I}$}{
  $\mathbf{F}_\text{inp} = \textsc{{UNet}}\left(\mathbf{I}\right)$; \\
  $\mathbf{P} = \mathbf{P}_{ini}$;
  
  \For{$t$ times}{
    $\mathbf{F}_\text{rend} = \textsc{{DeepTextureRender}}\left(\mathbf{P}, \mathcal{V}, \mathcal{F}, \mathcal{C}\right)$;  \\
    $\mathbf{e} = \text{vec}(\mathbf{F}_\text{inp}) - \text{vec}(\mathbf{F}_\text{rend}$);  \\
    $\mathbf{J} = \textsc{{Jacobian}}\left(\mathbf{F}_\text{rend}, \mathbf{P}, \mathcal{V}\right)$; \\
    $\Delta \mathbf{P} = \textsc{{PoseUpdate}}\left(\mathbf{e}, \mathbf{J}, \mathbf{P}\right)$; \\
    $\mathbf{P} = \mathbf{P} + \Delta \mathbf{P}$; \# Update Pose \\
  }
  $\mathbf{P}_{ref} = \mathbf{P}$; \\
  $\mathcal{L} = \textsc{{Loss}}\left(\mathbf{P}_\text{ref}, \mathbf{P}_\text{gt}, \mathbf{V} \right)$; \\
  $\textsc{UpdateParameters}(\mathcal{L}, \mathcal{C}, \textsc{UNet)}$; \\
}}
\end{algorithm}

\section{Experiments}
\label{sec:experiments}

\subsection{Implementation Details}
We train our model using Adam optimizer ~\cite{DBLP:journals/corr/KingmaB14} with a learning rate of $1 \times 10^{-3}$, decayed by $0.5$ every $100$ epochs. The number of channels $d$ in $\mathbf{F}_\text{inp}$ and $\mathbf{F}_\text{rend}$ is set to $3$ using grid search, and iterations $t$ in LM optimization is set to $5$.
We used pretrained PVNet~\cite{peng2019pvnet} on the LineMOD and Occlusion LineMOD datasets, and PoseCNN~\cite{xiang2018posecnn} on the YCB-Video~\cite{xiang2018posecnn} dataset as the initial pose estimator $\Omega$. The encoder of U-Net~\cite{RFB15a} consisting of ResNet-18~\cite{7780459} shares its weights with the PVNet, and PoseCNN and only the weights of the decoder are trained. Therefore, RePOSE simply can reuse the deep features extracted from the initial pose estimator, which reduces the computational cost. Following ~\cite{peng2019pvnet}, we also add $500$ synthetic and fused images for LineMOD and $20$K synthetic images for YCB-Video to avoid overfitting during training. In accordance with the convention, to evaluate the scores on the Occlusion LineMOD dataset, we use the model trained by using only the LineMOD dataset.

\subsection{Datasets}
All experiments are performed on the LineMOD~\cite{linemod}, Occlusion LineMOD~\cite{10.1007/978-3-319-10605-2_35}, and YCB-Video~\cite{xiang2018posecnn} datasets. The LineMOD dataset contains images of small texture-less objects in a cluttered scene under different illumination. High-quality template 3D models of the objects in the images are also provided for render and compare based pose estimation. The Occlusion LineMOD dataset is a subset of the LineMOD dataset focused mainly on the occluded objects. YCB-Video~\cite{xiang2018posecnn} dataset contains images of objects from the YCB-object set~\cite{7251504}. We use ADD (-S)~\cite{linemod} and AUC of ADD(-S) scores as our evaluation metrics.

\begin{table*}[t]
\caption{Results on the YCB-Video dataset using \textit{RGB only}. The results for DeepIM~\cite{li2018deepim} are computed using the official pre-trained model, and the score inside the parentheses are the reported results from the paper. Refinement FPS denotes FPS of running only a pose refinement network. RePOSE w/ track includes the runtime for CNN feature extraction of a real image. FPS is reported with refinement of 5 objects.}
\centering
\scalebox{0.9} {
\begin{tabular}{c||cccccc||ccc|ccc}
\hline
Metric  & PoseCNN~\cite{xiang2018posecnn} & \multicolumn{2}{c}{DeepIM~\cite{li2018deepim}} & PVNet~\cite{peng2019pvnet} & \multicolumn{2}{c}{CosyPose~\cite{labbe2020}} & \multicolumn{3}{c}{RePOSE} & \multicolumn{3}{c}{RePOSE w/ track} \\  \hline
AUC, ADD(-S) & 61.3 & 74.0 & 75.5 (81.9) & 73.4 & 84.1 & \textbf{84.5} & 70.5 & 79.4 & 80.8 & 70.1 & 80.6 & 82.0 \\
AUC, ADD-S & 75.2 & 83.1 & 83.1 (88.1) & - & \textbf{89.8} & \textbf{89.8} & 80.4 & 85.9 & 86.7 & 79.9 & 87.2 & 88.5 \\
ADD(-S) & 21.3 & 43.2 & 53.6 & - & 74.3 & \textbf{75.6} & 41.7 & 58.9 & 60.3 & 40.2 & 61.6 & 62.1 \\ \hline
Refinement FPS  & - & 22 & 6 & - & 26 & 13 & \textbf{181} & 111 & 80 & 125 & 90 & 71 \\ \hline
\#Iterations & - & 1 & 4 & - & 1 & 2 & 1 & 3 & 5 & 1 & 3 & 5 \\
\hline
\end{tabular}
}
\label{tab:ycb}
\end{table*}

\subsection{Evaluation Metrics}

\paragraph{ADD(-S) score.}
\label{sec:ADDS}
ADD(-S) score~\cite{linemod,xiang2018posecnn} is a standard metric which calculates the average distance between objects transformed by the predicted pose $\mathbf{\hat{P}} = \{ \mathbf{\hat{R}},  \mathbf{\hat{t}} \}$, and the ground-truth pose $\mathbf{P}_\text{gt} = \{ \mathbf{R}_\text{gt},  \mathbf{t}_\text{gt} \}$ using  vertices $\mathbf{V}_i$ of the template 3D model $\mathcal{M}$. The distance is calculated as follows;

\begin{equation}
  \frac{1}{N} \sum_i^N ||\left(\mathbf{\hat{R}} \mathbf{V}_i + \mathbf{\hat{t}} \right) - \left(\mathbf{R}_{gt} \mathbf{V}_i + \mathbf{t}_{gt} \right)||
\end{equation}
For symmetric objects such as \texttt{eggbox} and \texttt{glue}, we use the following distance metric,
\begin{equation}
  \frac{1}{N} \sum_i^N \min_{0 \le j \le N} ||\left(\mathbf{\hat{R}} \mathbf{V}_i + \mathbf{\hat{t}} \right) - \left(\mathbf{R}_{gt} \mathbf{V}_j + \mathbf{t}_{gt} \right)||
  \label{eq:add_s}
\end{equation}
The predicted pose is considered correct if this distance is smaller than $10$\% of the target object's diameter.
AUC of ADD(-S) computes the area under the curve of the distance used in ADD(-S). The pose predictions with distance larger than $0.1$m are not included in computing the AUC. We use AUC of ADD(-S) to evaluate the performance on the YCB-Video dataset~\cite{xiang2018posecnn}.

\begin{table}[t]
\small
\centering
\caption{Comparison of RePOSE on Linemod dataset with recent methods including PVNet~\cite{peng2019pvnet}, DPOD~\cite{Zakharov_2019_ICCV}, HybridPose~\cite{Song_2020_CVPR}, and EfficientPose~\cite{bukschat2020efficientpose} using the ADD(-S) score. \# of wins denotes in how many objects the method achieves the best score.}
\scalebox{0.8}{
\begin{tabular}{c||cccc|c}
  \hline
  Object      & PVNet & DPOD& HybridPose & EfficientPose & RePOSE \\ \hline
  Ape         & 43.6 & 87.7          & 63.1         & \textbf{89.4} & 79.5 \\
  Benchvise   & 99.9 & 98.5          & 99.9         & 99.7          & \textbf{100}  \\
  Camera      & 86.9 & 96.1          & 90.4         & 98.5          & \textbf{99.2} \\
  Can         & 95.5 & 99.7          & 98.5         & 99.7          & \textbf{99.8} \\
  Cat         & 79.3 & 94.7          & 89.4         & 96.2          & \textbf{97.9} \\
  Driller     & 96.4 & 98.8          & 98.5         & \textbf{99.5} & 99.0 \\
  Duck        & 52.6 & 86.3          & 65.0         & \textbf{89.2} & 80.3 \\
  Eggbox      & 99.2 & 99.9          & \textbf{100} & \textbf{100}  & \textbf{100}  \\
  Glue        & 95.7 & 98.7          & 98.8         & \textbf{100}  & 98.3 \\
  Holepuncher & 81.9 & 86.9          & 89.7         & 95.7          & \textbf{96.9} \\
  Iron        & 98.9 & \textbf{100}  & \textbf{100} & 99.1          & \textbf{100}  \\
  Lamp        & 99.3 & 96.8          & 99.5         & \textbf{100}  &  99.8 \\
  Phone       & 92.4 & 94.7          & 94.9         & 98.5          & \textbf{98.9} \\ \hline
  Average     & 86.3 & 95.2          & 91.3         & \textbf{97.4} & 96.1 \\ \hline
  \# of wins  & 0    & 1             & 2            & 6             & \textbf{8} \\ \hline
\end{tabular}
}
\label{tab:result_of_linemod}
\end{table}

\begin{table}[t]
\small
\centering
\caption{Comparison of RePOSE on Occlusion LineMOD dataset with recent methods including PVNet~\cite{peng2019pvnet}, DPOD~\cite{Zakharov_2019_ICCV}, 
and HybridPose~\cite{Song_2020_CVPR} using the ADD(-S) score. Note, we exclude EfficientPose~\cite{bukschat2020efficientpose} as it is trained on the Occlusion LineMOD dataset. \# of wins denotes in how many objects the method achieves the best score.}
\label{tab:result_of_occlusion_linemod}
\scalebox{0.94}{
\begin{tabular}{c||ccc|c}
  \hline
  Object      & PVNet & DPOD & HybridPose & RePOSE \\ \hline
  Ape         & 15.8 & - & 20.9          & \textbf{31.1} \\
  Can         & 63.3 & - & 75.3          & \textbf{80.0} \\
  Cat         & 16.7 & - & 24.9          & \textbf{25.6} \\
  Driller     & 65.7 & - & 70.2          & \textbf{73.1} \\
  Duck        & 25.2 & - & 27.9          & \textbf{43.0} \\
  Eggbox      & 50.2 & - & \textbf{52.4} & 51.7 \\
  Glue        & 49.6 & - & 53.8          & \textbf{54.3} \\
  Holepuncher & 39.7 & - & \textbf{54.2} & 53.6         \\ \hline
  Average     & 40.8 & 47.3 & 47.5    & \textbf{51.6} \\ \hline
  \# of wins  & 0    & -    & 2 & \textbf{6} \\ \hline
\end{tabular}
}
\end{table}

\subsection{Quantitative Evaluations}

\paragraph{Results on the LineMOD and Occlusion LineMOD datasets.}
As shown in~\Cref{tab:result_of_linemod,tab:result_of_occlusion_linemod}, RePOSE achieves the state of the art ADD(-S) scores on the Occlusion LineMOD dataset. In comparison to PVNet~\cite{peng2019pvnet}, RePOSE successfully refines the initial pose estimate in all the objects, achieving an improvement of $9.8$\% and $10.8$\% on the LineMOD and Occlusion LineMOD dataset respectively. On the LineMOD dataset, our score is comparable to the state-of-the art EfficientPose~\cite{bukschat2020efficientpose}. The key difference is mainly on \texttt{ape} and \texttt{duck} where our initial pose estimator PVNet~\cite{peng2019pvnet} performs poorly. Interestingly, for small objects like \texttt{ape} and \texttt{duck} in the Occlusion LineMOD dataset, we show a significant improvement of $10.2$ and $15.1$ respectively over the prior art HybridPose~\cite{Song_2020_CVPR}.

\paragraph{Results on the YCB-Video dataset.}
\Cref{tab:ycb} shows the result on the YCB-Video dataset~\cite{xiang2018posecnn}. We also performed experiments using RePOSE as a 6D object tracker using the tracking algorithm proposed in \cite{li2018deepim}. RePOSE achieves comparable performance with other methods with a $4$ times faster runtime of $80$ FPS for refinement of 5 objects. Further, the result with tracking demonstrates that RePOSE is useful as a real-time 6D object tracker. Note, the scores are heavily affected by the use and amount of synthetic data and various data augmentation~\cite{labbe2020}. For instance, CosyPose~\cite{labbe2020} used one million synthetic images during training, making it hard to compare against fairly. However, our method achieves comparable performance using $500$ times less training images.

\subsection{Ablation Study}
All ablations for RePOSE are conducted on the LineMOD and Occlusion LineMOD datasets using PVNet~\cite{peng2019pvnet} as an initial pose estimator. We report the results in \Cref{tab:result_of_linemod_ablation,tab:result_of_occlusion_linemod_ablation}.

\begin{table}[t]
\small
\caption{Ablation study of feature representation, feature warping, and a refinement network on the LineMOD dataset. RGB denotes pose refinement using photometric error. FW denotes feature warping after extraction from a CNN or deep texture rendering following first iteration. DPOD denotes using DPOD's refinement network and PVNet as an initial pose estimator. FW, DPOD, and RePOSE are trained with the same dataset, we report the ADD(-S) scores.}
\centering
\scalebox{0.7} {
\begin{tabular}{c||c|ccc|cc}
  \hline
  Object     & PVNet~\cite{peng2019pvnet} & RGB & CNN w/ FW & DPOD & Ours w/ FW & Ours \\ \hline
  Ape        & 43.6 & 5.81 & 65.4 & 51.2 & 75.9          & \textbf{79.5} \\
  Benchvise  & 99.9 & 75.6 & 99.8 & 99.5 & \textbf{100}  & \textbf{100}  \\
  Camera     & 86.9 & 7.06 & 96.3 & 91.1 & 98.2          & \textbf{99.2} \\
  Can        & 95.5 & 3.05 & 99.1 & 95.7 & 99.4          & \textbf{99.8} \\
  Cat        & 79.3 & 3.00 & 88.6 & 92.4 & 92.7          & \textbf{97.9} \\
  Driller    & 96.4 & 80.9 & 7.6 & 98.2 & 98.7          & \textbf{99} \\
  Duck       & 52.6 & 0.00 & 76.2 & 71.3 & \textbf{84.6} & 80.3 \\
  Eggbox     & 99.2 & 8.64 & 96.4 & 99.9 & \textbf{100}  & \textbf{100}  \\
  Glue       & 95.7 & 5.40 & 97.2 & 97.6 & 98.2          & \textbf{98.3} \\
  Holepuncher& 81.9 & 18.7 & 77.2 & 89.7 & 95.1          & \textbf{96.9} \\
  Iron       & 98.9 & 40.7 & 98.7 & 97.9 & 99.7          & \textbf{100}  \\
  Lamp       & 99.3 & 34.9 & 91.8 & 95.5 & \text{100}   & 99.8 \\
  Phone     & 92.4 & 14.6 & 94.9 & 97.2 & 98.7          & \textbf{98.9} \\ \hline
  Average    & 86.3 & 23.0 & 90.7 & 90.5 & 95.5          & \textbf{96.1} \\ \hline
\end{tabular}
}
\label{tab:result_of_linemod_ablation}
\end{table}

\begin{table}[t]
\small
\caption{Ablation study of feature representation, feature warping, and a refinement network on the Occlusion LineMOD dataset. We report the ADD(-S) scores, all other details are same as in \Cref{tab:result_of_linemod_ablation}.}
\centering
\scalebox{0.72} {
\begin{tabular}{c||c|ccc|cc}
  \hline
  Object     & PVNet~\cite{peng2019pvnet} & RGB & CNN w/ FW & DPOD & Our w/ FW & Ours \\ \hline
  Ape        & 15.8 & 4.96 & 22.7 & 22.0 & 25.8 & \textbf{31.1} \\
  Can        & 63.3 & 5.22 & 66.4 & 71.1 & 61.3 & \textbf{80.0} \\
  Cat        & 16.7 & 0.17 & 11.7 & 21.9 & 19.4 & \textbf{25.6} \\
  Driller    & 65.7 & 61.7 & 72.1 & 68.3 & 71.1 & \textbf{73.1} \\
  Duck       & 25.2 & 1.80 & 36.5 & 30.8 & 40.8 & \textbf{43.0} \\
  Eggbox     & 50.2 & 7.75 & 45.4 & 42.4 & 47.7 & \textbf{51.7} \\
  Glue       & 49.6 & 1.88 & 45.6 & 41.3 & 49.4 & \textbf{54.3} \\
  Holepuncher& 39.7 & 21.5 & 40.8 & 43.3 & 40.2 & \textbf{53.6} \\ \hline
  Average    & 40.8 & 13.1 & 42.9 & 42.6 & 44.5 & \textbf{51.6} \\ \hline
\end{tabular}
}
\label{tab:result_of_occlusion_linemod_ablation}
\end{table}

\begin{figure}[t]
\centering
\begin{tabular}{ccc}
  \begin{minipage}{0.3\linewidth}
    \includegraphics[scale=0.23]{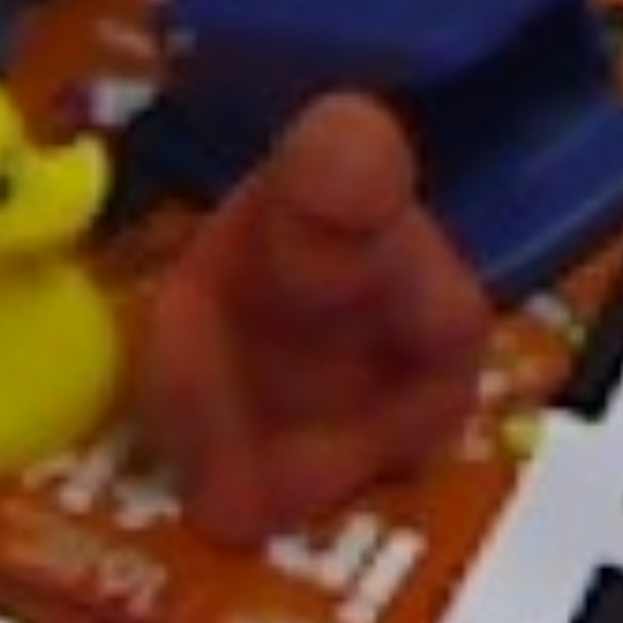}
  \end{minipage} &
  \begin{minipage}{0.3\linewidth}
    \includegraphics[scale=0.23]{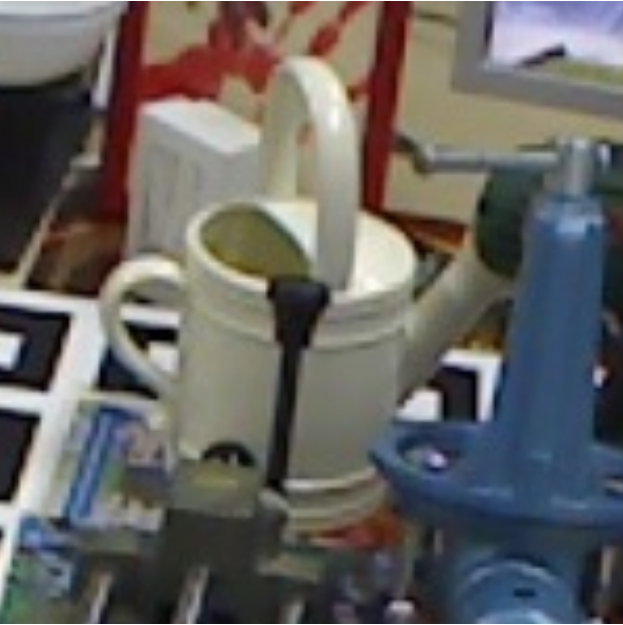}
  \end{minipage} &
  \begin{minipage}{0.3\linewidth}
    \includegraphics[scale=0.23]{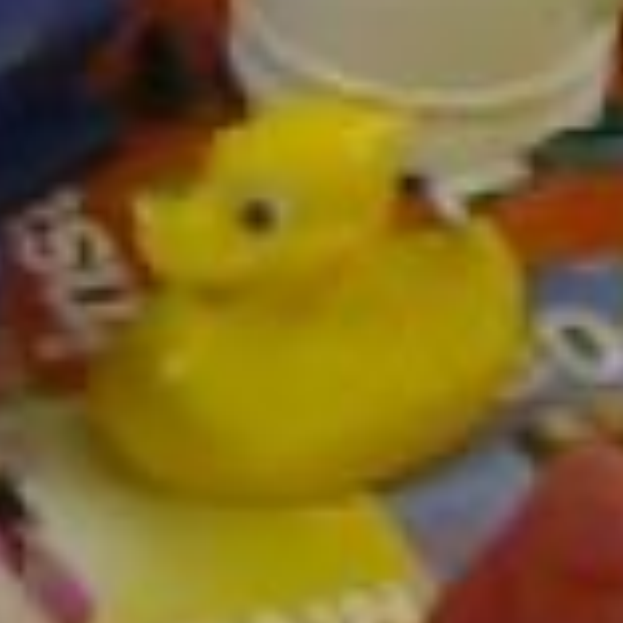}
  \end{minipage} \\ \\

  \begin{minipage}{0.3\linewidth}
    \includegraphics[scale=0.23]{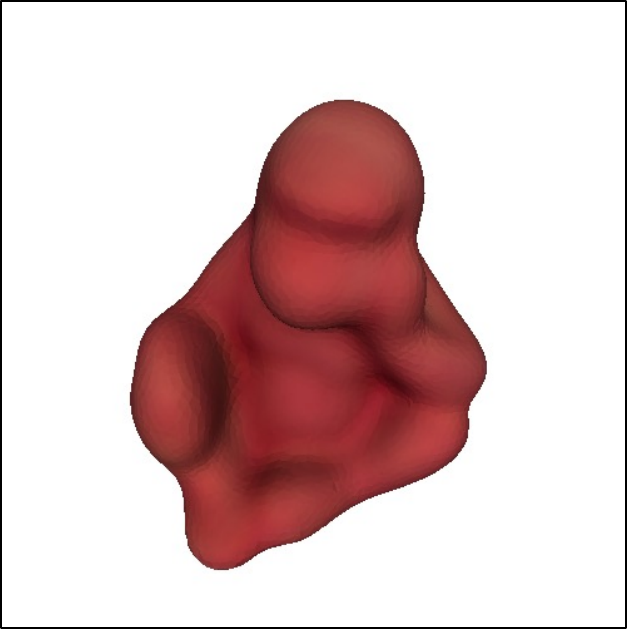}
    \caption*{(a) Ape}
  \end{minipage} &
  \begin{minipage}{0.3\linewidth}
    \includegraphics[scale=0.23]{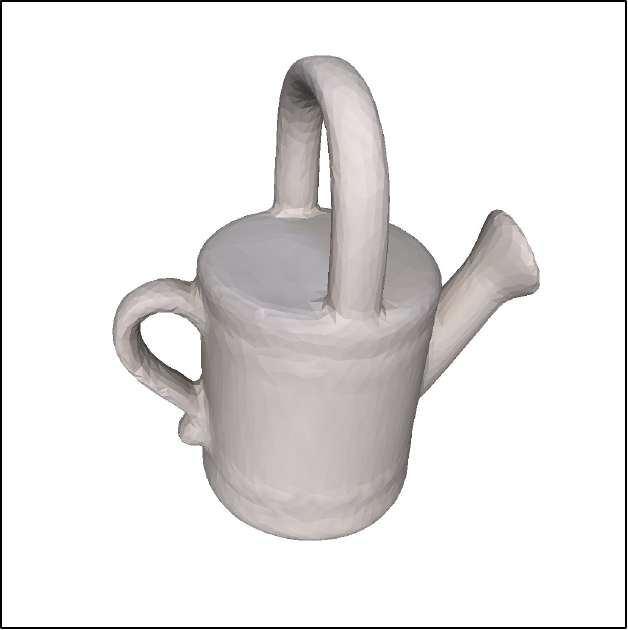}
    \caption*{(b) Can} 
  \end{minipage} &
  \begin{minipage}{0.3\linewidth}
    \includegraphics[scale=0.23]{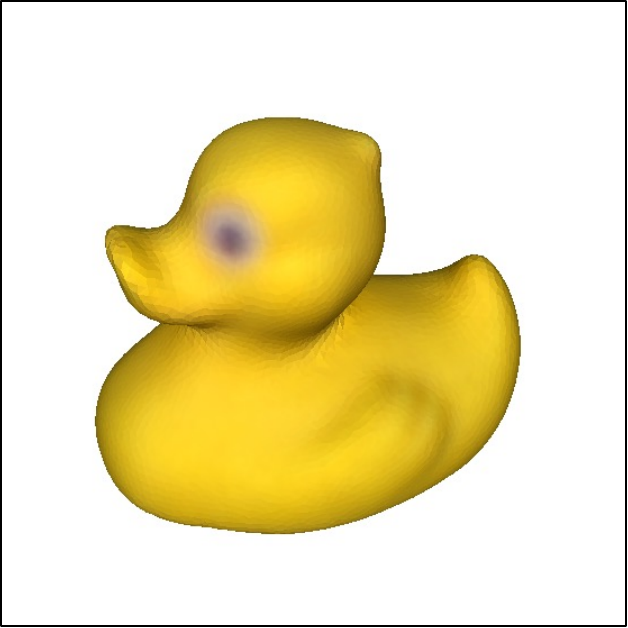}
    \caption*{(c) Duck}
  \end{minipage}
\end{tabular}
\caption{Comparison of object's appearance between an input RGB image and rendered image. Difference of illumination makes pose refinement in RGB space challenging. Furthermore, RGB images may have the region with the same color as the object. This background noise becomes an obstacle in terms of convergence properties. These texture-less objects make it challenging to compute the image gradient which is essential to optimize a pose.}
\label{fig:rgb_vs_orig}
\end{figure}

\paragraph{RGB vs Deep Texture.}
Instead of using learnable deep textures $\mathcal{C}$, we perform experiments using an original RGB image and rendered image with scanned colors. The inference is all the same except we are using photometric error between two images.
The experimental result reported in  \Cref{tab:result_of_linemod_ablation,tab:result_of_occlusion_linemod_ablation} show that the ADD(-S) score drops significantly after optimization in all the objects using RGB representation. As illustrated in \Cref{fig:rgb_vs_orig}, the LineMOD dataset has three main challenges which makes the pose refinement using the photometric error difficult --- 1) Illumination changes between the input RGB image and synthetic rendering, 2) Poor image gradients due to texture-less objects, 3) Background confusion \ie the background color is similar to the object's color.
The ADD(-S) scores drop largely due to these key reasons.
On the contrary, RePOSE with learnable deep textures is able to converge within few iterations because of the robustness of deep textures to the above challenges.
\Cref{tab:result_of_linemod_ablation,tab:result_of_occlusion_linemod_ablation} clearly demonstrate the effectiveness of our learnable deep textures over using scanned colors for the template 3D model.

\paragraph{CNN with Feature Warping vs Feature Rendering.}
Feature warping (FW) is commonly used to minimize photometric or feature-metric error through a non-linear least squares such as Gauss-Newton or Levenberg-Marquardt method~\cite{10.1007/978-3-642-15552-9_3,Triggs:1999:BAM:646271.685629}.
We conduct an experiment to compare a CNN with feature warping and our proposed feature rendering using the deep texture renderer.
In a CNN with feature warping, $\mathbf{F}_\text{rend}$ is extracted in the same fashion as the $\mathbf{F}_\text{inp}$ using a CNN on a normalized synthetic rendering of the template 3D model. This is done just once, following which the feature is warped based on the updated pose at each iteration.
The result is shown in \Cref{tab:result_of_linemod_ablation,tab:result_of_occlusion_linemod_ablation}.
On the LineMOD dataset, we observed on average small improvments by the feature warping. The ADD(-S) score only allows the pose estimator to have an mean vertex distance error of $10$\% of the object's diameter. In this task, this means only $2$ to $3$ pixel displacement error in 2D image space is allowed especially for small objects. However, it is challenging to train a CNN to extract features with accurate image gradients required for fine-grained pose refinement.
On the contrary, our deep texture renderer can compute accurate gradients as the neighborhood vertices on the template 3D model are not strongly correlated.
This local constraint is critical for fast and accurate pose refinement.

Furthermore, we perform additional experiments to verify the effect of feature warping.
To this end, we warp the feature extracted by deep texture renderer based on the updated pose (Ours w/ FW).
The result in \Cref{tab:result_of_linemod_ablation} shows that Ours w/ FW achieves 9.2\% absolute improvement from PVNet~\cite{peng2019pvnet} on the LineMOD dataset~\cite{linemod}.
However, \Cref{tab:result_of_occlusion_linemod_ablation} demonstrates the limited ability on the Occlusion LineMOD dataset~\cite{10.1007/978-3-319-10605-2_35}.
From this result, we figure out that warping has an inferior influence on refinement of occluded objects.
We conjecture that this difference comes from the fact that warping can not deal with large pose error because unlike our proposed RePOSE, feature warping can only consider the visible surface at the first step.
Being different from the methods using feature warping, our iterative deep texture rendering method can generate a feature with a complete shape.
We believe this characteristics of feature rendering leads to successful pose refinement.

\paragraph{Comparison with the latest refinement network on the LineMOD dataset.}
We compare our refinement network with the latest fully CNN-based refinement network proposed in the paper of DPOD~\cite{Zakharov_2019_ICCV}.
In this experiment, we use the same initial pose estimator~\cite{peng2019pvnet}.
Since DPOD is fully CNN-based, we increased the amount of the dataset by twice.
The refinement network of DPOD outputs a refined pose based on a cropped input RGB image and a synthetic rendering with an initial pose estimate.
The experimental result in \Cref{tab:result_of_linemod_ablation,tab:result_of_occlusion_linemod_ablation} shows DPOD fails to refine pose well when trained with the small amount of the dataset.
The refinement network of DPOD estimates a refined pose directly and do not consider projective geometry explicitly.
This means their network needs to learn not only deep features but also mapping of the deep feature into an object's pose from training data.
Several papers~\cite{brachmann2016differentiable,Sattler_2019_CVPR,Brachmann_2018_CVPR,Sattler2017EfficientE} report that learning a less complex task can achieve better accuracy and generalization in a 6D camera localization task.
Also, we assume the low ADD(-S) score on Occlusion LineMOD dataset implies its low generalization performance to occluded objects.
Our network only trains deep features and a refined object's pose is acquired by solving minimization problem based on projective geometry.
From this experimental result, we believe the same principle proposed in the field of 6D camera localization is still valid in 6D object pose estimation.

\begin{table}[t]
\caption{Comparison of number of iterations and refinement runtime. ADD(-S) on the Occlusion LineMOD dataset is reported in this table. Our proposed network is trained by using a pose loss for 5 iterations.}
\centering
\scalebox{1.0} {
\begin{tabular}{c|c||c|c}
  \hline
  Method & Iteration & ADD(-S) Score & Runtime \\ \hline
  AAE~\cite{Zakharov_2019_ICCV} & - & - & 200 ms \\
  SSD6D~\cite{Zakharov_2019_ICCV} & - & - & 24 ms \\
  DPOD~\cite{Zakharov_2019_ICCV} & - & 47.3 & 5 ms \\
  \hline
  \multirow{5}{*}{Ours}
       & 0 & 40.8 & 0 ms  \\
       & 1 & 45.7 & 4.1 ms \\ 
       & 2 & 48.6 & 5.8 ms \\ 
       & 3 & 50.1 & 7.5 ms \\ 
       & 4 & 51.0 & 9.2 ms \\
       & 5 & 51.6 & 10.9 ms \\ \hline
\end{tabular}
}
\label{tab:iteration_analysis}
\\\centering
\vspace*{1mm}
\end{table}

\paragraph{Number of iteration and run time analysis.}
Our proposed refinement network, RePOSE can adjust the trade-off between the accuracy and run time by changing the number of iterations.
We show the ADD(-S) score and the run time on the Occlusion LineMOD dataset with each iteration count in \Cref{tab:iteration_analysis}.
On a machine equipped with Nvidia RTX2080 Super GPU and Ryzen 7 3700X CPU, our method takes $1.7$ ms per iteration (deep texture rendering + pose update through LM optimization \cite{10.1007/BFb0067700}). This result shows our method achieves higher performance with the faster or comparable runtime than prior art.

\section{Conclusion}
Real-time pose estimation needs accurate and fast pose refinement. Our proposed method, RePOSE uses efficient deep texture renderer to perform pose refinement at $92$ FPS and has practical applications as a real-time 6D object tracker. Our experiments show that learnable deep textures coupled with the efficient non-linear optimization results in accurate 6D object poses. Further, our ablations highlight the fundamental limitations of a convolutional neural network to extract critical information useful for pose refinement. We believe that the concept of using efficient renderers with learnable deep textures instead of a CNN for pose refinement is an important conceptual change and will inspire a new research direction for real-time 6D object pose estimation.

\section{Acknowledgment}
This work is funded in part by the Department of Homeland Security award 2017-DN-077-ER0001, JST AIP Acceleration, Grant Number JPMJCR20U1, and JST CREST, Grant Number JPMJCR19F5, Japan.

\bibliographystyle{ieee_fullname}
\bibliography{arxiv}

\begin{thebibliography}{10}\itemsep=-1pt

\bibitem{EinScanPro}
Einscan pro 2x: https://www.einscan.com/handheld-3d-scanner/einscan-pro-2x/.

\bibitem{Triggs:1999:BAM:646271.685629}
Bundle adjustment - a modern synthesis.
\newblock In {\em Proceedings of the International Workshop on Vision
  Algorithms: Theory and Practice}, 2000.

\bibitem{10.1007/978-3-642-15552-9_3}
Sameer Agarwal, Noah Snavely, Steven~M. Seitz, and Richard Szeliski.
\newblock Bundle adjustment in the large.
\newblock In {\em ECCV}, 2010.

\bibitem{Alismail-2016-5532}
Hatem~Said Alismail, Brett Browning, and Simon Lucey.
\newblock Photometric bundle adjustment for vision-based slam.
\newblock In {\em ACCV}, 2016.

\bibitem{10.1023/B:VISI.0000011205.11775.fd}
Simon Baker and Iain Matthews.
\newblock Lucas-kanade 20 years on: A unifying framework.
\newblock In {\em IJCV}, 2004.

\bibitem{10.1007/978-3-319-10605-2_35}
Eric Brachmann, Alexander Krull, Frank Michel, Stefan Gumhold, Jamie Shotton,
  and Carsten Rother.
\newblock Learning 6d object pose estimation using 3d object coordinates.
\newblock In {\em ECCV}, 2014.

\bibitem{brachmann2016differentiable}
Eric Brachmann, Alexander Krull, Sebastian Nowozin, Jamie Shotton, Frank
  Michel, Stefan Gumhold, and Carsten Rother.
\newblock {DSAC - Differentiable RANSAC for Camera Localization}.
\newblock In {\em CVPR}, 2016.

\bibitem{Brachmann_2018_CVPR}
Eric Brachmann and Carsten Rother.
\newblock Learning less is more - 6d camera localization via 3d surface
  regression.
\newblock In {\em CVPR}, 2018.

\bibitem{bukschat2020efficientpose}
Yannick Bukschat and Marcus Vetter.
\newblock Efficientpose: An efficient, accurate and scalable end-to-end 6d
  multi object pose estimation approach.
\newblock In {\em CoRR}, 2020.

\bibitem{7251504}
Berk Calli, Arjun Singh, Aaron Walsman, Siddhartha Srinivasa, Pieter Abbeel,
  and Aaron~M. Dollar.
\newblock The ycb object and model set: Towards common benchmarks for
  manipulation research.
\newblock In {\em ICAR}, 2015.

\bibitem{BPnP2020}
Bo Chen, Alvaro Parra, Nan Cao, and Tat-Jun Chin.
\newblock End-to-end learnable geometric vision by backpropagating pnp
  optimization.
\newblock In {\em CVPR}, 2020.

\bibitem{Clark_2018_ECCV}
Ronald Clark, Michael Bloesch, Jan Czarnowski, Stefan Leutenegger, and
  Andrew~J. Davison.
\newblock Learning to solve nonlinear least squares for monocular stereo.
\newblock In {\em ECCV}, 2018.

\bibitem{10.1007/BFb0054760}
T.~F. Cootes, G.~J. Edwards, and C.~J. Taylor.
\newblock Active appearance models.
\newblock In {\em ECCV}, 1998.

\bibitem{compact_rotation_derivative}
Guillermo Gallego and Anthony Yezzi.
\newblock A compact formula for the derivative of a 3-d rotation in exponential
  coordinates.
\newblock {\em Journal of Mathematical Imaging and Vision}, 2015.

\bibitem{7780459}
Kaiming He, Xiangyu Zhang, Shaoqing Ren, and Jian Sun.
\newblock Deep residual learning for image recognition.
\newblock In {\em CVPR}, 2016.

\bibitem{linemod}
Stefan Hinterstoisser, Vincent Lepetit, Slobodan Ilic, Stefan Holzer, Gary
  Bradski, Kurt Konolige, and Nassir Navab.
\newblock Model based training, detection and pose estimation of texture-less
  3d objects in heavily cluttered scenes.
\newblock In {\em ACCV}, 2012.

\bibitem{kato2018renderer}
Hiroharu Kato, Yoshitaka Ushiku, and Tatsuya Harada.
\newblock Neural 3d mesh renderer.
\newblock In {\em CVPR}, 2018.

\bibitem{Kehl_2017_ICCV}
Wadim Kehl, Fabian Manhardt, Federico Tombari, Slobodan Ilic, and Nassir Navab.
\newblock Ssd-6d: Making rgb-based 3d detection and 6d pose estimation great
  again.
\newblock In {\em ICCV}, 2017.

\bibitem{DBLP:journals/corr/KingmaB14}
Diederik~P. Kingma and Jimmy Ba.
\newblock {Adam: {A} Method for Stochastic Optimization}.
\newblock In {\em ICLR}, 2015.

\bibitem{labbe2020}
Y. {Labbe}, J. {Carpentier}, M. {Aubry}, and J. {Sivic}.
\newblock Cosypose: Consistent multi-view multi-object 6d pose estimation.
\newblock In {\em ECCV}, 2020.

\bibitem{10.1007/s11263-008-0152-6}
Vincent Lepetit, Francesc Moreno-Noguer, and Pascal Fua.
\newblock Epnp: An accurate o(n) solution to the pnp problem.
\newblock {\em IJCV}, 2009.

\bibitem{li2018deepim}
Yi Li, Gu Wang, Xiangyang Ji, Yu Xiang, and Dieter Fox.
\newblock Deepim: Deep iterative matching for 6d pose estimation.
\newblock In {\em ECCV}, 2018.

\bibitem{10.5555/1623264.1623280}
Bruce~D. Lucas and Takeo Kanade.
\newblock {An Iterative Image Registration Technique with an Application to
  Stereo Vision}.
\newblock In {\em IJCAI}, 1981.

\bibitem{Matthews-2003-8630}
Iain Matthews and Simon Baker.
\newblock Active appearance models revisited.
\newblock Number CMU-RI-TR-03-02, Pittsburgh, PA, 2003.

\bibitem{10.1007/BFb0067700}
Jorge~J. Mor{\'e}.
\newblock The levenberg-marquardt algorithm: Implementation and theory.
\newblock In {\em Numerical Analysis}, 1978.

\bibitem{7219438}
R. {Mur-Artal}, J.~M.~M. {Montiel}, and J.~D. {Tardós}.
\newblock {ORB-SLAM: A Versatile and Accurate Monocular SLAM System}.
\newblock In {\em T-RO}, 2015.

\bibitem{Oberweger_2018_ECCV}
Markus Oberweger, Mahdi Rad, and Vincent Lepetit.
\newblock Making deep heatmaps robust to partial occlusions for 3d object pose
  estimation.
\newblock In {\em ECCV}, 2018.

\bibitem{peng2019pvnet}
Sida Peng, Xiaowei Liu, and Hujun Bao.
\newblock Pvnet: Pixel-wise voting network for 6dof pose estimation.
\newblock In {\em CVPR}, 2019.

\bibitem{DBLP:journals/corr/RadL17}
Mahdi Rad and Vincent Lepetit.
\newblock Bb8: A scalable, accurate, robust to partial occlusion method for
  predicting the 3d poses of challenging objects without using depth.
\newblock In {\em ICCV}, 2017.

\bibitem{RFB15a}
O. Ronneberger, P.Fischer, and T. Brox.
\newblock {U-Net: Convolutional Networks for Biomedical Image Segmentation}.
\newblock In {\em MICCAI}, 2015.

\bibitem{Sattler2017EfficientE}
Torsten Sattler, B. Leibe, and L. Kobbelt.
\newblock Efficient \& effective prioritized matching for large-scale
  image-based localization.
\newblock {\em PAMI}, 2017.

\bibitem{Sattler_2019_CVPR}
Torsten Sattler, Qunjie Zhou, Marc Pollefeys, and Laura Leal-Taixe.
\newblock Understanding the limitations of cnn-based absolute camera pose
  regression.
\newblock In {\em CVPR}, 2019.

\bibitem{7780814}
J.~L. {Schönberger} and J. {Frahm}.
\newblock Structure-from-motion revisited.
\newblock In {\em CVPR}, 2016.

\bibitem{Simonyan15}
Karen Simonyan and Andrew Zisserman.
\newblock Very deep convolutional networks for large-scale image recognition.
\newblock In {\em ICLR}, 2015.

\bibitem{Song_2020_CVPR}
Jiaru Song and Qixing Huang.
\newblock Hybridpose: 6d object pose estimation under hybrid representations.
\newblock In {\em CVPR}, 2020.

\bibitem{Sundermeyer_2018_ECCV}
Martin Sundermeyer, Zoltan-Csaba Marton, Maximilian Durner, Manuel Brucker, and
  Rudolph Triebel.
\newblock Implicit 3d orientation learning for 6d object detection from rgb
  images.
\newblock In {\em ECCV}, 2018.

\bibitem{Tang2019}
Chengzhou Tang and Ping Tan.
\newblock {BA-Net: Dense Bundle Adjustment Network}.
\newblock {\em ICLR}, 2019.

\bibitem{tremblay2018corl:dope}
Jonathan Tremblay, Thang To, Balakumar Sundaralingam, Yu Xiang, Dieter Fox, and
  Stan Birchfield.
\newblock {Deep Object Pose Estimation for Semantic Robotic Grasping of
  Household Objects}.
\newblock In {\em CoRL}, 2018.

\bibitem{gnnet}
L. {von Stumberg}, P. {Wenzel}, Q. {Khan}, and D. {Cremers}.
\newblock {GN-Net: The Gauss-Newton Loss for Multi-Weather Relocalization}.
\newblock In {\em ICRA}, 2020.

\bibitem{wang2021nemo}
Angtian Wang, Adam Kortylewski, and Alan Yuille.
\newblock Nemo: Neural mesh models of contrastive features for robust 3d pose
  estimation.
\newblock In {\em ICLR}, 2021.

\bibitem{xiang2018posecnn}
Yu Xiang, Tanner Schmidt, Venkatraman Narayanan, and Dieter Fox.
\newblock {PoseCNN: A Convolutional Neural Network for 6D Object Pose
  Estimation in Cluttered Scenes}.
\newblock In {\em RSS}, 2018.

\bibitem{Zakharov_2019_ICCV}
Sergey Zakharov, Ivan Shugurov, and Slobodan Ilic.
\newblock Dpod: 6d pose object detector and refiner.
\newblock In {\em ICCV}, 2019.

\bibitem{Zhang2014}
Zhengyou Zhang.
\newblock {Iterative Closest Point (ICP)}.
\newblock In {\em Computer Vision: A Reference Guide}, 2014.

\end{thebibliography}

\clearpage

\appendix

\begin{table*}
\caption{Comparison of the median of absolute angular and relative translation error on the LindMOD dataset~\cite{linemod}. We do not report the rotation error of symmetric objects (eggbox, and glue) because of its non-unique rotation representation.}
\centering
\vspace*{1mm}
\begin{tabular}{c||cc|cc|cc|cc}
  \hline
  {Object}
             & \multicolumn{2}{c|}{PVNet~\cite{peng2019pvnet}} &
               \multicolumn{2}{c|}{CNN w/ FW} &
               \multicolumn{2}{c|}{Ours w/ FW} &
               \multicolumn{2}{c}{Ours} \\ \hline
  & Rotation & Translation & Rotation & Translation & Rotation & Translation & Rotation & Translation\\ \hline
  Ape        & $2.213^{\circ} $ & 0.119 & $1.849^{\circ}$ & 0.069 & $1.446^{\circ}$ & 0.055 & $1.197^{\circ}$ & 0.051 \\
  Benchvise  & $1.030^{\circ} $ & 0.022 & $0.857^{\circ}$ & 0.022 & $0.644^{\circ}$ & 0.014 & $0.757^{\circ}$ & 0.010 \\
  Cam        & $1.183^{\circ} $ & 0.045 & $0.896^{\circ}$ & 0.030 & $1.322^{\circ}$ & 0.073 & $0.713^{\circ}$ & 0.023 \\
  Can        & $0.958^{\circ} $ & 0.032 & $1.238^{\circ}$ & 0.033 & $0.995^{\circ}$ & 0.040 & $0.674^{\circ}$ & 0.017 \\
  Cat        & $1.260^{\circ} $ & 0.050 & $1.177^{\circ}$ & 0.041 & $0.941^{\circ}$ & 0.036 & $0.857^{\circ}$ & 0.035 \\
  Driller    & $1.008^{\circ} $ & 0.029 & $0.812^{\circ}$ & 0.023 & $1.057^{\circ}$ & 0.060 & $0.773^{\circ}$ & 0.014 \\
  Duck       & $1.701^{\circ} $ & 0.078 & $1.701^{\circ}$ & 0.055 & $1.531^{\circ}$ & 0.042 & $1.481^{\circ}$ & 0.053 \\
  Eggbox     & -                & 0.056 & -               & 0.074 & -               & 0.048 & -               & 0.025 \\
  Glue       & -                & 0.050 & -               & 0.049 & -               & 0.040 & -               & 0.036 \\
  Holepuncher& $1.265^{\circ} $ & 0.052 & $1.548^{\circ}$ & 0.058 & $1.295^{\circ}$ & 0.060 & $1.009^{\circ}$ & 0.029 \\
  Iron       & $1.205^{\circ} $ & 0.027 & $1.223^{\circ}$ & 0.027 & $0.927^{\circ}$ & 0.020 & $0.911^{\circ}$ & 0.015 \\ 
  Lamp       & $1.050^{\circ} $ & 0.029 & $1.235^{\circ}$ & 0.042 & $1.054^{\circ}$ & 0.019 & $0.902^{\circ}$ & 0.020 \\ 
  Phone      & $1.208^{\circ} $ & 0.040 & $1.122^{\circ}$ & 0.032 & $0.925^{\circ}$ & 0.019 & $0.889^{\circ}$ & 0.025 \\ \hline
  Average    & $1.280^{\circ} $ & 0.048 & $1.242^{\circ}$ & 0.043 & $1.103^{\circ}$ & 0.040 & $0.924^{\circ}$ & 0.027 \\ \hline
\end{tabular}
\label{tab:rot_trans_ablation}
\end{table*}

\begin{table*}
\caption{Comparison of the median of absolute angular and relative translation error on the Occlusion LindMOD dataset~\cite{10.1007/978-3-319-10605-2_35}. We do not report the rotation error of symmetric objects (eggbox, and glue) because of its non-unique rotation representation.}
\centering
\vspace*{1mm}
\begin{tabular}{c||cc|cc|cc|cc}
  \hline
  {Object}
             & \multicolumn{2}{c|}{PVNet~\cite{peng2019pvnet}} &
               \multicolumn{2}{c|}{CNN w/ FW} &
               \multicolumn{2}{c|}{Ours w/ FW} &
               \multicolumn{2}{c}{Ours} \\ \hline
  & Rotation & Translation & Rotation & Translation & Rotation & Translation & Rotation & Translation\\ \hline
  Ape        & $4.103^{\circ}$ & 0.210 & $4.222^{\circ}$ & 0.185 & $4.015^{\circ}$ & 0.171 & $3.871^{\circ}$ & 0.157 \\
  Can        & $2.722^{\circ}$ & 0.068 & $2.879^{\circ}$ & 0.069 & $2.793^{\circ}$ & 0.067 & $2.704^{\circ}$ & 0.043 \\
  Cat        & $6.012^{\circ}$ & 0.239 & $6.480^{\circ}$ & 0.363 & $6.552^{\circ}$ & 0.335 & $5.852^{\circ}$ & 0.274 \\
  Driller    & $2.774^{\circ}$ & 0.072 & $2.567^{\circ}$ & 0.131 & $2.762^{\circ}$ & 0.120 & $2.545^{\circ}$ & 0.056 \\
  Duck       & $6.923^{\circ}$ & 0.156 & $6.795^{\circ}$ & 0.115 & $6.537^{\circ}$ & 0.108 & $6.533^{\circ}$ & 0.102 \\
  Eggbox     & -               & 0.325 & -               & 0.295 & -               & 0.284 & -               & 0.318 \\
  Glue       & -               & 0.275 & -               & 0.246 & -               & 0.235 & -               & 0.266 \\
  Holepuncher& $3.969^{\circ}$ & 0.121 & $3.917^{\circ}$ & 0.116 & $3.918^{\circ}$ & 0.126 & $3.629^{\circ}$ & 0.088 \\ \hline
  Average    & $4.417^{\circ}$ & 0.183 & $4.477^{\circ}$ & 0.190 & $4.430^{\circ}$ & 0.181 & $4.189^{\circ}$ & 0.163 \\ \hline
\end{tabular}
\label{tab:rot_trans_ablation_occ}
\end{table*}

\section{Implementation Details}
RePOSE uses the initial estimated pose of PVNet on the LineMOD and Occlusion LineMOD dataset, and of PoseCNN on the YCB-Video dataset. PoseCNN has VGG-16~\cite{Simonyan15} as the backbone network. However, since RePOSE reuses the deep feature from PoseCNN, its slow runtime of VGG-16 can be problematic in the case of tracking. Instead, we use ResNet-18 as the backbone network of PoseCNN. RePOSE makes use of the deep feature from ResNet-18, however, we still rely on the initial pose of the original PoseCNN with VGG-16. For tracking, the U-Net decoder of RePOSE also outputs the segmentation mask and use it as an additional signal to detect whether an object is lost or not. RePOSE learns its parameters separately per object on the LineMOD and Occlusion LineMOD dataset. In contrast to that, RePOSE learns the parameters for all the objects simultaneously on the YCB-Video dataset. We use \texttt{pvnet-rendering}\footnote{https://github.com/zju3dv/pvnet-rendering} to generate synthetic images. However, we may be able to improve accuracy if images generated by photorealistic rendering are used for training. The neural textures, which are the outputs of the MLP that takes learnable parameters as input (ref. Fig.1 of the main paper), are trained along with the MLP and input learnable parameters.

\section{Derivation of Camera Jacobian Matrix}

\paragraph{Notation:}
$\mathbf{R} \in \text{SO}(3)$ denotes a rotation matrix, $\mathbf{t} \in \mathbb{R}^{3}$ denotes a translation vector, $\mathbf{w} \in \mathfrak{so}(3)$ denotes a rotation vector, $\mathbf{P} = \begin{pmatrix}
  \mathbf{w} & \mathbf{t}
\end{pmatrix} \in \mathbb{R}^{6}$ is a pose representation using a rotation and translation vector, $f_x$ and $f_y$ are focal lengths, $p_x$ and $p_y$ are the principal point offsets, $\mathbf{X}$ and $\mathbf{x}$ denote a homogeneous and inhomogeneous 2D image coordinate, subscript $w, c$ denotes a coordinate is defined in the world, and camera coordinate system respectively, and $\mathbf{Id} \in \mathbb{R}^{3 \times 3}$ is an $3 \times 3$ identity matrix.
$\mathbf{R}$ and $\mathbf{w}$ represents the same rotation.

\paragraph{Derivation:}
Jacobian matrix $\mathbf{J}$ of the objective function with respect to a pose $\mathbf{P}$ is required to perform deep feature-based pose optimization.
We show the detailed derivation of a camera jacobian matrix $\frac{\partial \mathbf{x}}{\partial \mathbf{P}}$ at each pixel in Equation 9.
The camera jacobian matrix can be decomposed more as follows;

\begin{equation}
  \frac{\partial \mathbf{x}}{\partial \mathbf{P}} = \begin{pmatrix}
    \cfrac{\partial \mathbf{R}}{\partial \mathbf{w}}
      \cfrac{\partial \mathbf{x}}{\partial \mathbf{R}}\\[3ex]
      \cfrac{\partial \mathbf{x}}{\partial \mathbf{t}}
    \end{pmatrix} \in \mathbb{R}^{6 \times 2}
\end{equation}
Using the derivative calculation method of a rotation matrix with respect to a rotation vector proposed in \cite{compact_rotation_derivative}, the following equation is acquired.

\begin{equation}
\begin{split}
   \cfrac{\partial \mathbf{R}}{\partial \mathbf{w}} &=
     \begin{pmatrix}
       \cfrac{\partial \mathbf{R}}{\partial {w}_1} &
       \cfrac{\partial \mathbf{R}}{\partial {w}_2} &
       \cfrac{\partial \mathbf{R}}{\partial {w}_3} \\
     \end{pmatrix}^{T} \\
     &=
     \begin{pmatrix}
       \textit{vec} \left( \cfrac{w_{1}[\mathbf{w}]_{\times}+\left[\mathbf{w} \times(\mathbf{Id}-\mathbf{R}) \mathbf{e}_{1}\right]_{\times}}{\|\mathbf{w}\|^{2}}\mathbf{R} \right) \\
       \textit{vec} \left( \cfrac{w_{2}[\mathbf{w}]_{\times}+\left[\mathbf{w} \times(\mathbf{Id}-\mathbf{R}) \mathbf{e}_{2}\right]_{\times}}{\|\mathbf{w}\|^{2}}\mathbf{R} \right) \\
       \textit{vec} \left( \cfrac{w_{3}[\mathbf{w}]_{\times}+\left[\mathbf{w} \times(\mathbf{Id}-\mathbf{R}) \mathbf{e}_{3}\right]_{\times}}{\|\mathbf{w}\|^{2}}\mathbf{R} \right) \\
     \end{pmatrix} \in \mathbb{R}^{3 \times 9}
\end{split}
\end{equation}
where \textit{vec} is a vectorize operation, $[\mathbf{x}]_{\times}$ is a conversion from a 3-d vector to a skew-symmetric matrix, $\mathbf{e}_{i}$ is a $i$th 3-d basis vector.
$\mathbf{R}$ is regarded as a 9-d vector for simplicity.
A camera and image coordinate $\mathbf{x}_c$ and $\mathbf{x}$ can be calculated as follows;

\begin{equation}
  \mathbf{x}_c = \mathbf{R} \mathbf{x}_{w} + \mathbf{t}
\label{eq:camera_projection}
\end{equation}

\begin{equation}
    \mathbf{x} = \begin{pmatrix}
                   \cfrac{f_x x_c}{z_c} + p_x \\
                   \cfrac{f_y y_c}{z_c} + p_y \\
                \end{pmatrix}
    \label{eq:image_coordinate}
\end{equation}
Using \Cref{eq:camera_projection,eq:image_coordinate}, the following derivatives can be obtained.

\begin{equation}
   \cfrac{\partial \mathbf{x}}{\partial \mathbf{R}} =
     \begin{pmatrix}
       \cfrac{f_x x_w}{z_c} & 0 \\ \cfrac{f_x y_w}{z_c} & 0 \\ \cfrac{f_x z_w}{z_c} & 0 \\
       0 & \cfrac{f_y x_w}{z_c} \\ 0 & \cfrac{f_y y_w}{z_c} \\ 0 & \cfrac{f_y z_w}{z_c} \\
       - \cfrac{f_x x_c x_w}{z_c^2} & - \cfrac{f_y y_c x_w}{z_c^2} \\
       - \cfrac{f_x x_c y_w}{z_c^2} & - \cfrac{f_y y_c y_w}{z_c^2} \\
       - \cfrac{f_x x_c z_w}{z_c^2} & - \cfrac{f_y y_c z_w}{z_c^2} \\
     \end{pmatrix} \in \mathbb{R}^{9 \times 2}
\end{equation}

\begin{equation}
   \cfrac{\partial \mathbf{x}}{\partial \mathbf{t}} =
     \begin{pmatrix}
       \cfrac{f_x}{z_c} & 0 \\
       0 & \cfrac{f_y}{z_c}  \\
       - \cfrac{f_x x_c}{z_c^2} & - \cfrac{f_y y_c}{z_c^2} \\
     \end{pmatrix} \in \mathbb{R}^{3 \times 2}
\end{equation}

\section{Additional Ablation Study}

\paragraph{Ablation on effect of warping.}
We show the median of absolute angular and relative translation error on the LineMOD and Occlusion LineMOD datasets in \Cref{tab:rot_trans_ablation,tab:rot_trans_ablation_occ}.
The relative translation error is computed with respect to object diameter.
The initial pose error on the LineMOD dataset~\cite{linemod} is relatively small and the methods using warping improve score properly.
On the contrary, the initial pose error is large on the Occlusion LineMOD dataset.
In that case, feature warping becomes less effective.
Being different from the methods using feature warping, our iterative deep feature rendering method can generate a feature with a complete shape.
We believe this characteristics of feature rendering leads to successful reduction of the error of rotation and translation on both datasets.

\begin{table}[t]
    \centering
    \caption{Ablation of the channel size on the LineMOD dataset}
    \begin{tabular}{c|ccccc}
        \hline
        Channel Size & 1 & 3 & 5 & 7 & 9 \\ \hline
        ADD(-S) & 95.2 & \textbf{96.1} & 95.7 & 94.9 & 94.5 \\ \hline
    \end{tabular}
    \label{tab:ab_ch}
\end{table}

\begin{table}[t]
 \centering
 \caption{Runtime comparison among recent methods. In CNN with feature warping (FW) the initial feature is obtained using CNN and then warp the feature based on an updated pose. We assume the number of iterations is 5 for RePOSE and FW denotes feature warping. The FPS is reported with refinement of 5 objects.}
 \vspace{0.2cm}
 \label{tab:comp_strat}
 \begin{tabular}{|c||c|}
   \hline
   Method & Runtime \\ \hline
   DeepIM~\cite{li2018deepim} (1 iters) & 45.8 ms          \\
   DeepIM~\cite{li2018deepim} (4 iters) & 166.8 ms          \\
   CosyPose~\cite{labbe2020} (1 iters) & 38.3 ms          \\
   CosyPose~\cite{labbe2020} (2 iters) & 77.1 ms          \\ \hline
   \multicolumn{2}{|c|}{RePOSE w/ different feature extraction methods}           \\ \hline
   CNN w/ FW   & 19.6 ms         \\
   Ours w/ FW    & 13.5 ms         \\
   Ours & 12.4 ms \\ \hline
 \end{tabular}
\end{table}

\paragraph{Ablation on the number of channels in the neural textures}
We vary the number of channels in the neural textures. We report the results on the LineMOD dataset in Table \ref{tab:ab_ch}. Note, the results are comparable, however, setting number of channels to $3$ results in the best performance.

\begin{figure}[t]
  \centering
  \hspace*{-0.4cm}
  \includegraphics[scale=0.57]{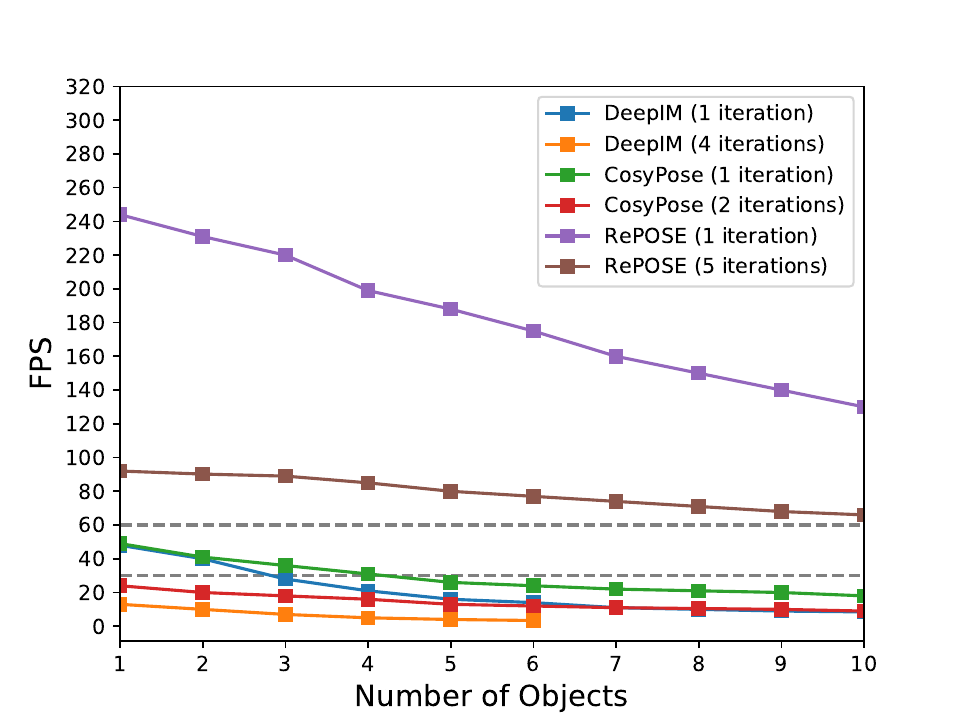}
  \caption{Trade-off between FPS and number of objects of recent methods.}
  \label{fig:runtime}
\end{figure}

\paragraph{Runtime Ablation on Feature Warping.}
DeepIM~\cite{li2018deepim} and CosyPose~\cite{labbe2020} run a CNN every iteration on a concatenated image of a zoomed input and rendered images to compare these two images and output a pose directly.
According to the ablation study by \cite{li2018deepim}, high-resolution zoomed-in is a key and it improves the ADD(-S) score by 23.4. 
However, as shown in \Cref{tab:comp_strat}, extracting image features from zoomed images multiple times leads to a slow runtime.
Instead, RePOSE runs a CNN once for an input image with the original resolution. Additionally, an image representation of a rendered image can be extracted within 1ms because of deep texture rendering. This makes the runtime of RePOSE faster than prior methods while keeping a comparable accuracy to the prior methods.
We also measure the runtime of RePOSE using different feature extraction methods for a rendered image.
As shown in \Cref{tab:comp_strat} in the supplemental and Table 4 and 5 in the main paper, RePOSE with deep texture rendering (Ours) achieves the fastest and highest accuracy among these three variants.

\paragraph{Runtime Ablation on the number of objects}
We investigate the trade-off between FPS and number of objects.
As shown in \ref{fig:runtime}, RePOSE is the only method which runs at over than 60 FPS even with refinement of 10 objects. 6D pose refinement is always performed after initial pose estimation. Thus, it is crucial to make sure it runs at faster than real-time (30 FPS). DeepIM causes an out of memory error when the number of object is more than 6 with a NVIDIA RTX2080 which has 8GB memory.

\end{document}